\documentclass[11pt,a4paper]{article}%
\usepackage[T1]{fontenc}
\usepackage[ttscale=0.875]{libertine}
\usepackage[table]{xcolor}
\usepackage[american]{babel}
\usepackage{xfrac}
\usepackage{bbm}
\usepackage{authblk}
\usepackage{amsthm,amssymb}
\usepackage{bm}

\usepackage[format=plain,font=it]{caption}
\usepackage{graphicx}
\usepackage{amsmath}
\usepackage{amsthm}
\usepackage{booktabs}
\usepackage{algorithm}
\usepackage{algorithmic}
\usepackage[switch]{lineno}
\usepackage{pdflscape}
\usepackage{arydshln}

\usepackage{geometry}
\usepackage[round]{natbib} 
\usepackage{mathtools} 
\usepackage{booktabs} 
\usepackage{tikz} 
\usetikzlibrary{positioning,arrows.meta}
\usepackage{multirow}

\usepackage{float}
\usepackage{subfig}
\usepackage{amsmath}
\usepackage{amsfonts}
\usepackage{comment}
\usepackage{dsfont}

\usepackage{array}
\newcolumntype{C}[1]{>{\centering\arraybackslash}m{#1}}
\newcolumntype{R}[1]{>{\raggedleft\arraybackslash}m{#1}}
\newcolumntype{L}[1]{>{\raggedright\arraybackslash}m{#1}}

\title{Optimal Transport on Categorical Data \\ for Counterfactuals using Compositional Data \\ and Dirichlet Transport\thanks{Agathe Fernandes Machado acknowledges that the project leading to this publication has received funding from OBVIA. Arthur Charpentier acknowledges funding from the SCOR Foundation for Science and the National Sciences and Engineering Research Council (NSERC) for funding (RGPIN-2019-07077). Ewen Gallic acknowledges funding from the French government under the ``France 2030'' investment plan managed by the French National Research Agency (reference: ANR-17-EURE-0020) and from Excellence Initiative of Aix-Marseille University -- A*MIDEX.\\Replication codes and companion e-book: \href{https://github.com/fer-agathe/transport-simplex}{https://github.com/fer-agathe/transport-simplex}}}

\usepackage{graphicx,pstricks,psfrag}
\definecolor{bleu}{RGB}{0,101,189}
\definecolor{vert}{HTML}{004D40}
\definecolor{rose}{HTML}{D81B60}
\definecolor{bleuTOL}{HTML}{332288}
\definecolor{wongBlack}{RGB}{0,0,0}
\definecolor{wongGold}{RGB}{230, 159, 0}
\definecolor{wongLightBlue}{RGB}{86, 180, 233}
\definecolor{wongGreen}{RGB}{0, 158, 115}
\definecolor{wongYellow}{RGB}{240, 228, 66}
\definecolor{wongBlue}{RGB}{0, 114, 178}
\definecolor{wongOrange}{RGB}{213, 94, 0}
\definecolor{wongPurple}{RGB}{204, 121, 167}
\definecolor{colUncalibrated}{RGB}{191, 191, 191}
\definecolor{colRecalibrated}{RGB}{197, 214, 231}

\definecolor{bleuTOL}{HTML}{332288}
\definecolor{vertTOL}{HTML}{117733}
\definecolor{vertClairTOL}{HTML}{44AA99}
\definecolor{bleuClairTOL}{HTML}{88CCEE}
\definecolor{sableTOL}{HTML}{DDCC77}
\definecolor{parmeTOL}{HTML}{CC6677}
\definecolor{magentaTOL}{HTML}{AA4499}
\definecolor{roseTOL}{HTML}{882255}

\definecolor{wongPurple}{RGB}{204, 121, 167}
\definecolor{wongLightBlue}{RGB}{86, 180, 233}
\definecolor{gris}{HTML}{A9A9A9}

\usepackage{hyperref}
\hypersetup{
    plainpages=false,
    bookmarksopen,
    bookmarksnumbered,
    unicode=true,          
    pdftoolbar=true,        
    pdfmenubar=true,        
    pdffitwindow=false,     
    pdfstartview={FitH},    
    pdftitle={Optimal Transport on Categorical Data for Counterfactuals using Compositional Data and Dirichlet Transport},    
    pdfauthor={Fernandes Machado, Agathe and Charpentier, Arthur and Gallic, Ewen},     
    pdfkeywords={compositions, fairness, individual fairness, optimal transport, simplex, }, 
    pdfnewwindow=true,      
    colorlinks=true,       
    linkcolor=black,          
    citecolor={black},        
    filecolor={rose},      
    urlcolor={wongBlue},           
}

\usepackage{doi}
\author[1]{Agathe~Fernandes~Machado\thanks{Corresponding author: \href{mailto:fernandes_machado.agathe@courrier.uqam.ca}{fernandes\_machado.agathe@courrier.uqam.ca}}}
\author[1]{Arthur~Charpentier}
\author[2,3]{Ewen~Gallic}

\affil[1]{%
    \footnotesize Département de Mathématiques\\
    Université du Québec à Montréal\\
    Montréal, Québec, Canada
}
\affil[2]{%
    \footnotesize CNRS - Université de Montréal CRM -- CNRS
}
\affil[3]{%
    \footnotesize Aix Marseille Univ, CNRS, AMSE\\
    Marseille, France
}

\usepackage[misc]{ifsym}
\makeatletter
\def\@fnsymbol#1{%
   \ifcase#1\or
   \TextOrMath ~ \dagger\or
   \TextOrMath {\footnotesize\Letter} \dagger\or
   \TextOrMath \textdaggerdbl \ddagger \or
   \TextOrMath \textsection  \mathsection\or
   \TextOrMath \textparagraph \mathparagraph\or
   \TextOrMath \textbardbl \|\or
   \TextOrMath {\textdagger\textdagger}{\dagger\dagger}\or
   \TextOrMath {\textdaggerdbl\textdaggerdbl}{\ddagger\ddagger}\else
   \@ctrerr \fi
}
\makeatother

\usepackage{fancyhdr} 
\newcommand{\authornames}{\footnotesize\textsc{Fernandes Machado, Charpentier, Gallic}}
\usepackage{etoolbox}
\makeatletter
\patchcmd{\NAT@test}{\else \NAT@nm}{\else \NAT@nmfmt{\NAT@nm}}{}{}

\DeclareRobustCommand\citepos
  {\begingroup
   \let\NAT@nmfmt\NAT@posfmt
   \NAT@swafalse\let\NAT@ctype\z@\NAT@partrue
   \@ifstar{\NAT@fulltrue\NAT@citetp}{\NAT@fullfalse\NAT@citetp}}

\let\NAT@orig@nmfmt\NAT@nmfmt
\def\NAT@posfmt#1{\NAT@orig@nmfmt{#1's}}
\makeatother

\begin{document}

\maketitle

\begin{abstract}
Recently, optimal transport-based approaches have gained attention for deriving counterfactuals, e.g., to quantify algorithmic discrimination. However, in the general multivariate setting, these methods are often opaque and difficult to interpret. To address this, alternative methodologies have been proposed, using causal graphs combined with iterative quantile regressions \citep{plevcko2020fair} or sequential transport \citep{machado2024sequential}
to examine fairness at the individual level, often referred to as ``counterfactual fairness.'' Despite these advancements, transporting categorical variables remains a significant challenge in practical applications with real datasets.
In this paper, we propose a novel approach to address this issue. Our method involves (1) converting categorical variables into compositional data and (2) transporting these compositions within the probabilistic simplex of $\mathbb{R}^d$. We demonstrate the applicability and effectiveness of this approach through an illustration on real-world data, and discuss limitations.
\end{abstract}

\section{Introduction}

\subsection{Counterfactuals}

Counterfactual analysis, the third level in \cite{pearl2009causality}'s causal hierarchy, is widely used in machine learning, policy evaluation, economics and causal inference. It involves reasoning about ``what could have happened'' under alternative scenarios, providing insights into causality and decision-making effectiveness. An example could be the concept of counterfactual fairness, as introduced by \cite{Kusner17}, that ensures fairness by evaluating how decisions would change under alternative, counterfactual conditions. Counterfactual fairness focuses on mitigating bias by ensuring that sensitive attributes, such as race, gender, or socioeconomic status, do not unfairly influence outcomes.

In the counterfactual problem, we consider data $\{(s_i,\mathbf{x}_i),i=1,\cdots,n\}$, where $s$ denotes a binary ``treatment'' (taking values in $\{0,1\}$). With generic notations,  the counterfactual version of $(0,\mathbf{x})$ {can be constructed as} $(1,T^\star(\mathbf{x}))$, where $T^\star$ is the optimal {transport} (OT) mapping from $\mathbf{X}|S=0$ to $\mathbf{X}|S=1$, as discussed in \cite{black2020fliptest}, \cite{charpentier2023optimal} and \cite{de2024transport}. Unfortunately, this multivariate mapping is usually both complicated to estimate, and hard to interpret. If $\mathbf{x}$ is univariate, it is simply a quantile interpretation: if $x$ is associated to rank probability $u$ within group $s=0$, then its counterfactual version should be associated with the same rank probability in group $s=1$ (mathematically, $T^\star = F^{-1}_1\circ F_0$, where $F_j: \mathbb{R}\rightarrow[0,1], j=\{0,1\}$ denotes the cumulative distribution in group $j$, and $F_j^{-1}$ is the generalized inverse, i.e., the quantile function). In higher dimensions, one could consider multivariate quantiles, as in \cite{hallin2021distribution} or \cite{hallin2024multivariate}, but the heuristics is still hard to interpret.
While OT-based counterfactual methods have been proposed to assess counterfactual fairness \cite{black2020fliptest,de2024transport}, an alternative approach introduced by \cite{plevcko2020fair} is grounded in causal graphs (DAGs). In this framework, the outcome $y$ depends on variables $(s, \mathbf{x})$, where the sensitive attribute $s$ ``is a source'' (a vertex without parents) and $y$ is a ``sink'' (a vertex without outgoing edges). Recently, \cite{machado2024sequential} unified these approaches by introducing sequential transport aligned with the ``topological ordering'' of a DAG.

For example, to test whether a predictor $\widehat m(\mathbf{x})$ is gender‐neutral; let the sensitive attribute $s$ be gender (binary genders for simplicity); compare its output on a woman's features $\mathbf{x}$ with that on her \textit{mutatis mutandis} male counterpart. Unlike a \textit{ceteris paribus} change, which flips $s$ while holding all other features fixed, a \textit{mutatis mutandis} intervention also adjusts any $x_j$ causally influenced by $s$. Thus, if $x_1$ is height, the counterfactual of a 5'4" woman would not be a 5'4" man but, say, a 5'10" man, via an OT map. While OT handles continuous attributes naturally, categorical features (e.g. occupation or neighbourhood) lack a canonical distance. As a result, generating counterfactuals (e.g. the male counterpart of a female \textit{nurse}, or where a Black \textit{resident of X} would live if they were White) becomes particularly challenging.

\subsection{The Case of Categorical Variables}

For absolutely continuous variables, the approaches of \cite{plevcko2020fair,plevcko2021fairadapt} on the one hand (based on quantile regressions) and \cite{black2020fliptest,charpentier2023optimal,de2024transport,machado2024sequential} (based on OT) are quite similar.

If \cite{plevcko2020fair} considered quantile regressions for absolutely continuous variables, the case of ordered categorical variables is considered (at least with some sort of meaningful ordering) in the section related to ``Practical aspects and extensions.'' Discrete optimal transport between two marginal multinomial distributions is considered, but as discussed, it suffers multiple limitations. Here, we will consider an alternative approach, based on the idea of transforming categorical variables into continuous ones, coined ``compositional variables'' in \cite{chayes1971ratio}, and then, using ``Dirichlet optimal transport,'' on those compositions.

While motivated by counterfactual fairness, the primary aim of this study is to present the core of a method for deriving counterfactuals for categorical data, applicable to any context requiring counterfactual analysis. Here, for simplicity, we have set aside considerations related to the assumption of a known Structural Causal Model (SCM).\footnote{Details on how the method can be integrated within an SCM are discussed in Appendix~\ref{sec:appendix-scm}.}

\subsection{Agenda}

After recalling notations on OT in Section \ref{sec:ot}, we discuss how to transform categorical variables with $d$ categories into variables taking values in the simplex {$\mathcal{S}_d$ in $\mathbb{R}^d$}, i.e., compositional variables, in Section \ref{sec:categorical}. In Section~\ref{sec:topology}, we review the topological and geometrical properties of the probability simplex $\mathcal{S}_d \subset \mathbb{R}^d$. %
In Section~\ref{sec:representation}, we introduce the first methodology, which transports distributions within $\mathcal{S}_d$ via Gaussian OT. This approach relies on an alternative representation of probability vectors in the Euclidean space $\mathbb{R}^{d-1}$ and assumes approximate normality in the transformed space. %
In Section~\ref{sec:Sd:OT}, we present a second methodology, which operates directly on $\mathcal{S}_d$ using a tailored cost function instead of the standard quadratic cost. %
Theoretical aspects of this ``Dirichlet transport'' framework are discussed in Section~\ref{sec:theory}, while empirical strategies for matching categorical observations are developed in Section~\ref{subsec:matching}. %
Section~\ref{sec:application} provides two empirical illustrations using the \texttt{German Credit} and \texttt{Adult} datasets.

{Our main contributions can be summarized as follows:}
\begin{itemize}
    \item We propose a novel method to handle categorical variables in counterfactual modeling by using optimal transport directly on the simplex. This approach transforms categorical variables into compositional data, enabling the use of probabilistic representations that preserve the geometric structure of the simplex. 
    \item By integrating optimal transport techniques on this domain, the method ensures consistency with the properties of compositional data and offers a robust framework for counterfactual analysis in real-world scenarios.
    \item Our approach does not require imposing an arbitrary order on the labels of categorical variables.
\end{itemize}

\section{Optimal Transport}\label{sec:ot}

Given two metric spaces \(\mathcal{X}_0\) and \(\mathcal{X}_1\), consider a measurable map \(T:\mathcal{X}_0\to\mathcal{X}_1\) and a measure \(\mu_0\) on \(\mathcal{X}_0\).
The {push-forward} of \(\mu_0\) by \(T\) is the measure \(\mu_1 = T_{\#}\mu_0\) on \(\mathcal{X}_1\) defined by \(T_{\#}\mu_0(B)=\mu_0\big(T^{-1}({B})\big)\), \(\forall {B}\subset\mathcal{X}_1\).
For all measurable and bounded \(\varphi:\mathcal{X}_1\to\mathbb{R}\),
\[\int_{\mathcal{X}_1}\varphi (\mathbf{x}_1)T_{\#}\mu_0(\mathrm{d}\mathbf{x}_1) = 
\int_{\mathcal{X}_0}\varphi\big(T(\mathbf{x}_0)\big)\mu_0(\mathrm{d}\mathbf{x}_0).
\]
For our applications, if we consider measures \(\mathcal{X}_0=\mathcal{X}_1\) as a compact subset of \(\mathbb{R}^d\), then there exists \(T\) such that \(\mu_1 = T_{\#}\mu_0\), when \(\mu_0\) and \(\mu_1\) are two measures, and \(\mu_0\) is atomless, as shown in \cite{villani2003optimal} and \cite{santambrogio2015optimal}.
Out of those mappings from \(\mu_0\) to \(\mu_1\), we can be interested in ``optimal'' mappings, satisfying Monge problem, from \cite{monge1781memoire}, i.e., solutions of 
\begin{equation}\label{eq:transport:1}
\inf_{T_{\#}\mu_0=\mu_1} \int_{\mathcal{X}_0} c\big(\mathbf{x}_0,T(\mathbf{x}_0)\big)\mu_0(\mathrm{d}\mathbf{x}_0),
\end{equation}
for some positive ground cost function \(c:\mathcal{X}_0\times\mathcal{X}_1\to\mathbb{R}_+\).  In general settings, however, such a deterministic mapping \(T\) between probability distributions may not exist (in particular if \(\mu_0\) and \(\mu_1\) are not absolutely continuous, with respect to Lebesgue measure). This limitation motivates the Kantorovich relaxation of Monge's problem, as considered in \cite{kantorovich1942translocation},
\begin{equation}\label{eq:transport:2}
\inf_{\pi\in\Pi(\mu_0,\mu_1)} \int_{\mathcal{X}_0\times\mathcal{X}_1} c(\mathbf{x}_0,\mathbf{x}_1)\pi(\mathrm{d}\mathbf{x}_0,\mathrm{d}\mathbf{x}_1),
\end{equation}
with our cost function \(c\), where \({\displaystyle \Pi (\mu_0 ,\mu_1 )}\) is the set of all couplings of \(\mu_0\) and \(\mu_1\). This problem focuses on couplings rather than deterministic mappings It always admits solutions referred to as OT plans.
Observe that \(T^\star\) is an ``{increasing mapping},'' in the sense of being the gradient of a convex function, from \cite{brenier1991polar}). Finally, one should have in mind the the cost function $c$ is related to the geometry of sets $\mathcal{X}$.

\section{From Categorical to Compositional Data}\label{sec:categorical}

Using the notations of the introduction, consider a dataset $\{s,\mathbf{x}\}$ where features $\mathbf{x}$ are either numerical ({assumed to be} ``continuous''), or categorical. In the latter case, suppose that $\mathbf{x}_j$ takes values in $\{x_{j,1},\cdots,x_{j,d_j}\}$, or more conveniently, $[\![ d_j]\!]=\{1,\cdots,d_j\}$, corresponding to the $d_j$ categories (as in the standard ``One Hot" encoding).

The aim is to transform a categorical variable $x$, which takes values in $[\![ d]\!]$, into a numerical one in the simplex $\mathcal{S}_d$. To achieve this, we suggest using a probabilistic classifier. This classifier is based on the other features in $\mathbf{x}$, denoted by $\mathcal{X}_{-x}$. Mathematically, we consider a mapping from $\mathcal{X}_{-x}$ to $\mathcal{S}_d$ (and not to $[\![ d]\!]$ as in a standard multiclass classifier).
The most natural model for this transformation is the Multinomial Logistic Regression (MLR),  which is based on the ``softmax'' loss function. To normalize the output of the classifier into the simplex, we define the closure operator ${\mathcal {C}:\mathbb{R}_+^d\to\mathcal{S}_d}$ as
\[{\displaystyle {\mathcal {C}}[x_{1},x_{2},\dots ,x_{d}]=\left[{\frac {x_{1}}{\sum _{i=1}^{d}x_{i}}},{\frac {x_{2}}{\sum _{i=1}^{d}x_{i}}},\cdots ,{\frac {x_{d}}{\sum _{i=1}^{d}x_{i}}}\right], }\]
or shortly
\[
\mathcal {C}(\mathbf {x})=\frac{\mathbf {x}}{\mathbf {x}^\top\mathrm{1}},
\] 
where $\mathrm{1}$ is a vector of ones in $\mathbb{R}^d$. Then, in the MLR model, the transformation $\widehat{T}:\mathcal{X}_{-x}\to\mathcal{S}_d$ is given by
$$
\widehat{T}(\mathbf{x})= 
\mathcal {C}(1,e^{\mathbf {x}^\top\widehat{\boldsymbol{\beta}}_2},\cdots,e^{\mathbf {x}^\top\widehat{\boldsymbol{\beta}}_{d}})\in\mathcal{S}_d,
$$
where $\widehat{\boldsymbol{\beta}}_2, \dots, \widehat{\boldsymbol{\beta}}_d$ are the estimated coefficients for each category, and the first category is taken as the reference.
This procedure is described in Algorithm \ref{alg:0}.

\begin{algorithm}[tb]
\caption{From categorical variables into compositions.}\label{alg:0}
    \textbf{Input}: training dataset $\mathcal{D}=\{(s_i,\mathbf{x}_i)\}$\\
    \textbf{Input}: new observation $(s,\mathbf{x})$, with ${\mathbf{x}}_j$'s either in $\mathbb{R}$ or $[\![ d_j]\!]$\\
    \textbf{Output}: $(s,\tilde{\mathbf{x}})$, with $\tilde{\mathbf{x}}_j$'s either in $\mathbb{R}$ or $\mathcal{S}_{d_j}$\\
    \vspace{-.4cm}
\begin{algorithmic}
\FOR{\(j\in\{1,\cdots,k\}\)}
\IF{$\mathbf{x}_j\in[\![ d_j]\!]$}
\STATE estimate a MLR to predict categorical $\mathbf{x}_j$ using $\mathcal{D}$
\STATE get estimates $\widehat{\boldsymbol{\beta}}_2,\cdots,\widehat{\boldsymbol{\beta}}_{d_j}$ 
\STATE $\tilde{\mathbf{x}}_j\gets \mathcal {C}(1,e^{\mathbf {x}_j^\top\widehat{\boldsymbol{\beta}}_2},\cdots,e^{\mathbf {x}_j^\top\widehat{\boldsymbol{\beta}}_{d}})$
\ELSE 
\STATE $\tilde{\mathbf{x}}_j\gets\mathbf{x}_j$
\ENDIF\\
\ENDFOR\\
\end{algorithmic}
\end{algorithm}

{As an illustration, consider the \texttt{purpose} variable from the \texttt{German} dataset. For simplicity, this variable has been reduced to three categories: ${\text{C}, \text{E}, \text{O}}$ (representing cars, equipment, and other, respectively). More details on the dataset are provided in Section~\ref{sec:germancredit}. {The \texttt{purpose} variable is converted into a continuous variable using four models: (i) a GAM-MLR with splines for three continuous variables, (ii) a GAM-MLR incorporating these variables and seven categorical ones, (iii) a random forest, and (iv) a gradient boosting model.} Table~\ref{tab:4:model} presents the observed values in the first column for each model, along with the estimated scores for each category in the three remaining columns, corresponding to the transformed values ${T}^\star(\mathbf{x})$.}

Note that if we want to go back from compositions to categories, the standard approach is based on the majority (or argmax) rule.

\begin{table}[tb]
    \centering\setlength{\tabcolsep}{3.7pt}
    {\footnotesize 
    \begin{tabular}{cc}
    \begin{tabular}{|c|ccc|}
\multicolumn{4}{c}{GAM-MLR (1)}\\\hline
$x$ & ~~~~$\tilde{x}_{\text{C}}$& ~~~~$\tilde{x}_{\text{E}}$& ~~~~$\tilde{x}_{\text{O}}$  \\\hline
        E & 18.38\%&     {\bf 61.56\%}& 20.06\% \\ 
        C & 40.86\%&     {\bf 42.38\%}& 16.76\% \\ 
        E &  19.41\%&      {\bf 70.82\%}&   \phantom{8}9.77\%\\ 
        C & {\bf 47.04\%}&   26.83\%& 26.13\% \\\hline 
       \multicolumn{4}{c}{} \\
        \multicolumn{4}{c}{GAM-MLR (2)}\\\hline
 $x$&~~~~$\tilde{x}_{\text{C}}$& ~~~~$\tilde{x}_{\text{E}}$& ~~~~$\tilde{x}_{\text{O}}$  \\\hline
        E & \phantom{8}9.22\%&        {\bf 75.92\%}&    14.86\% \\ 
        C & {\bf 46.80\%}&        24.06\%&    29.14\% \\ 
        E & 11.23\%&        {\bf 79.07\%}&     \phantom{8}9.71\%\\ 
        C & {\bf 50.74\%}&        26.98\%&    22.28\% \\\hline 
        \end{tabular}&
    \begin{tabular}{|ccc|}
       \multicolumn{3}{c}{random forest}\\\hline
        ~~~~$\tilde{x}_{\text{C}}$& ~~~~$\tilde{x}_{\text{E}}$& ~~~~$\tilde{x}_{\text{O}}$ \\\hline
         23.68\%&     {\bf 46.32\%}& 30.00\%\\ 
         34.68\%&     {\bf 36.42\%}& 28.90\%\\ 
         16.87\%&     {\bf 76.51\%}&  \phantom{8}6.63\%\\ 
         {\bf 53.16\%}&     26.84\%& 20.00\% \\\hline
       \multicolumn{3}{c}{} \\
         \multicolumn{3}{c}{gradient boosting model}\\\hline
        ~~~~$\tilde{x}_{\text{C}}$& ~~~~$\tilde{x}_{\text{E}}$& ~~~~$\tilde{x}_{\text{O}}$ \\\hline
         11.25\%&       {\bf 68.51\%}&   20.24\% \\ 
         {\bf 61.14\%}&       13.10\%&   25.76\% \\ 
         12.48\%&       {\bf 75.58\%}&   11.94\%\\ 
         {\bf 51.12\%}&       25.17\%&   23.71\% \\\hline  
    \end{tabular} 
    \end{tabular}
    }
    \caption{Mappings from the \texttt{purpose} categorical variable $x$ to the compositional one $\tilde{\mathbf{x}}$, (in the \texttt{german credit} dataset), {for the first four individuals of the dataset}. The first two models are GAM-MLR (multinomial model with splines for continuous variables), then, a random forest, and a boosting algorithm.}
    \label{tab:4:model}
\end{table}

In the rest of the paper, given a dataset $\{s,\mathbf{x}\}$, all categorical variables are transformed into compositions, so that $\mathcal{X}$ is a product space of sets that are either $\mathbb{R}$ for numerical variables or $\mathcal{S}_d$ (type) for compositions ($d$ will change according to the number of categories).

In fact, for privacy issues, a classical strategy is to consider aggregated data on small groups (usually on a geographic level, per block, or per zip code), even if there is an ecological fallacy issue (that occurs when conclusions about individual behaviour or characteristics are incorrectly drawn based on aggregate data for a group, see \cite{king2004ecological}). Hence, using ``compositional data'' is quite natural in many cases, as unobserved categorical variables can often be represented as compositions {predicted from observed variables serving as proxies}. For example, in U.S. datasets,  racial information about individuals may not always be available. However, the proportions of groups such as ``White and European,'' ``Asian,'' ``Hispanic and Latino,'' ``Black or African American,'' etc., within a neighbourhood may be observed instead (see, e.g., \cite{cheng2010bayes}, \cite{naeini2015obtaining} and \cite{zadrozny2001} for more general discussions, or \cite{doi:10.1126/sciadv.adc9824} about the use of predicted probabilities when categories are not observed).

\section{Topology and Geometry of the Simplex}\label{sec:topology}

\begin{figure}[!htb]
    \centering
    \includegraphics[width=0.5\linewidth]{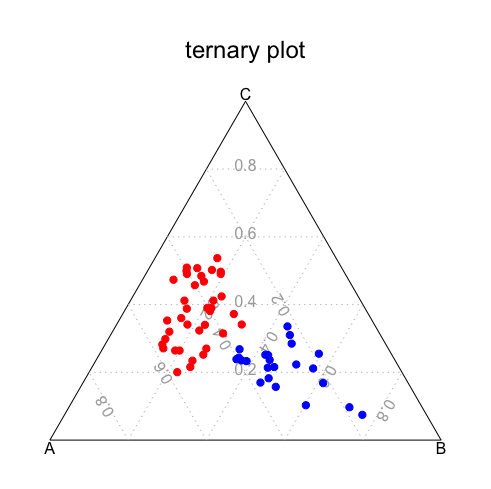}

{\small
\begin{tabular}{|l|p{0.065\linewidth}p{0.065\linewidth}p{0.075\linewidth}|}\hline
& ~~~A & ~~~B & ~~~C \\\hline
\textcolor{blue}{$\bullet$} & 29.831\% & 48.605\% & 21.564\% \\
\textcolor{red}{$\bullet$} & 43.713\% & 18.572\% & 37.715\% \\\hline 
\end{tabular}
}
    \caption{$n=61$ points in $\mathcal{S}_3$, with a toy dataset.}
    \label{fig:ternary1}
\end{figure}

The standard simplex of $\mathbb {R}^{d}$ is the regular polytope \({\displaystyle {\mathcal {S}}_{d}=\left\{\mathbf {x} \in \mathbb {R}_+^{d}\,\left|\,\mathbf {x}^\top\mathbf {1}=1\right.\right\}}\), but for convenience, consider the open version of that set,
\[{\displaystyle {\mathcal {S}}_{d}=\left\{\mathbf {x} \in (0,1)^{d}\,\left|\,\mathbf {x}^\top\mathbf {1}=1\right.\right\}.}\]
Following \cite{aitchison1982statistical}, define the inner product
\begin{equation}\label{eq:inner}
    {\displaystyle \langle \mathbf {x},\mathbf {y}\rangle ={\frac {1}{d}}\sum_{i<j}\log {\frac {x_{i}}{x_{j}}}\log {\frac {y_{i}}{y_{j}}}\qquad \forall x,y\in \mathcal{S}_{d}},
\end{equation}
and the simplex becomes a metric vector space if we consider the associated ``Aitchison distance,'' as coined in \cite{pawlowsky2001geometric}.
Figure \ref{fig:ternary1} shows $n=61$ points in $\mathcal{S}_3$. Each point $\mathbf{x}$ can be seen as a probability vector over $\{\text{A},\text{B},\text{C}\}$, drown either from a distribution $\mathbb{P}_0$ for red points or $\mathbb{P}_1$ for blue points. 

If we define the binary operator $\diamond$ on $\mathcal{S}_d$,
\[
\mathbf {x}\diamond \mathbf {y} = \left[{\frac {x_{1}y_1}{\sum _{i=1}^{d}x_{i}y_i}},\cdots ,{\frac {x_{d}y_d}{\sum _{i=1}^{d}x_{i}y_i}}\right],
\]
then $({\mathcal {S}_d},\diamond )$ is a commutative group, with identity element $d^{-1} \mathbf{1}$, and the inverse of $\mathbf {x}$ is 
\[\mathbf {x}^{-1}=
\left[{\frac {1/x_{1}}{\sum _{i=1}^{d}1/x_{i}}},\cdots ,{\frac {1/x_{d}}{\sum _{i=1}^{d}1/x_{i}}}\right]
=
\mathcal {C}(1/\mathbf {x}).\]

\section{Using an Alternative Representation of Simplex Data}\label{sec:representation}

A first strategy to define a transport mapping could be to use some isomorphism, $h:\mathcal{S}_{d}\rightarrow \mathcal{E}$ and then define the inverse mapping $h^{-1}:\mathcal{E}\rightarrow \mathcal{S}_{d}$, where $\mathcal{E}$ is some Euclidean space, classically $\mathbb{R}^{d-1}$, where the standard quadratic cost can be considered. This idea corresponds to the dual transport problem in \cite{pal2018exponentially}.

\subsection{Classical Transformations}\label{subsec:ilr:alr}

The additive log ratio (alr) transform is an isomorphism where \({\displaystyle \operatorname {alr} :\mathcal{S}_{d}\rightarrow \mathbb {R}^{d-1}}\), given by
\[{\displaystyle \operatorname {alr} (\mathbf{x})=\left[\log {\frac {x_{1}}{x_{d}}},\cdots ,\log {\frac {x_{d-1}}{x_{d}}}\right]}.
\]
Its inverse is, for any $\mathbf{z}\in\mathbb {R}^{d-1}$,
\[
     {\displaystyle \operatorname {alr}^{-1}(\mathbf{z})=\mathcal{C}(\exp(z_1),\cdots,\exp(z_{d-1}),1)=\mathcal{C}\big(\exp([\mathbf{z},0])\big)}.
\]
Such a map, from $\mathcal{S}_{d}$ to $\mathbb {R}^{d-1}$ is related to the so-called ``exponential coordinate system'' of the unit simplex, in \cite{pal2024difference}.
The center log ratio (clr) transform is both an isomorphism and an isometry where \({\displaystyle \operatorname {clr} :S^{d}\rightarrow \mathbb {R} ^{d}}\),
\[
     {\displaystyle \operatorname {clr} (\mathbf{x})=\left[\log {\frac {x_{1}}{\overline{\mathbf{x}}_g}},\cdots ,\log {\frac {x_{D}}{\overline{\mathbf{x}}_g}}\right]},
    \]
where $\overline{\mathbf{x}}_g$ denotes the geometric mean of $\mathbf{x}$. Observe that the inverse of this function is the softmax function, i.e.,
\[
     {\displaystyle \operatorname {clr}^{-1}(\mathbf{z})=\mathcal{C}(\exp(z_1),\cdots,\exp(z_d))=\mathcal{C}\big(\exp(\mathbf{z})\big)},~\mathbf{z}\in\mathbb {R} ^{d}.
    \]
Finally, the isometric log ratio (ilr) transform, defined in \cite{egozcue2003isometric}, is both an isomorphism and an isometry where \({\displaystyle \operatorname {ilr} :\mathcal{S}_{d}\rightarrow \mathbb{R} ^{d-1}}\), 
\[{\displaystyle \operatorname {ilr} (\mathbf{x})={\big [}\langle \mathbf{x},\Vec{e}_{1}\rangle ,\cdots ,\langle \mathbf{x},\Vec{e}_{d-1}\rangle {\big ]}}\]
for some orthonormal base $\{\Vec{e}_1,\cdots,\Vec{e}_{d-1},\Vec{e}_{d}\}$ of $\mathbb{R} ^{d}$. One can consider some matrix \(\mathbf{M}\), $d\times(d-1)$ such that \(\mathbf{M}\mathbf{M}^\top=\mathbb{I}_{d-1}\) and \(\mathbf{M}^\top\mathbf{M}=\mathbb{I}_{d}+\boldsymbol{1}_{d\times d}\). Then
\[{\displaystyle \operatorname {ilr} (\mathbf{x})=\operatorname {clr} (\mathbf{x})\mathbf{M}=\log(\mathbf{x})\mathbf{M} },\]
and
\[
     {\displaystyle \operatorname {ilr}^{-1}(\mathbf{z})=\mathcal{C}\big((\exp(\mathbf{z}\mathbf{M} ^\top)\big)},~\mathbf{z}\in\mathbb {R} ^{d-1}.
    \]

\begin{figure}[!htb]
    \centering
    \includegraphics[width=0.5\linewidth]{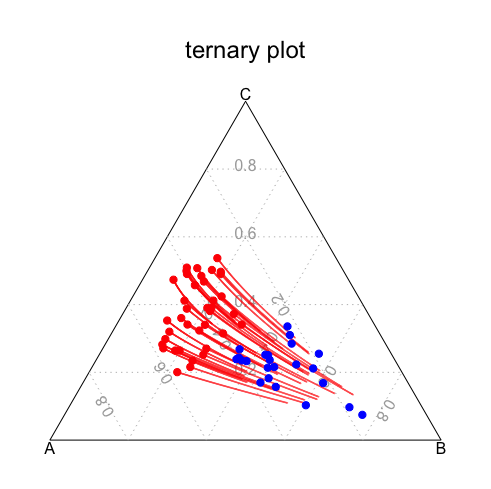}\includegraphics[width=0.5\linewidth]{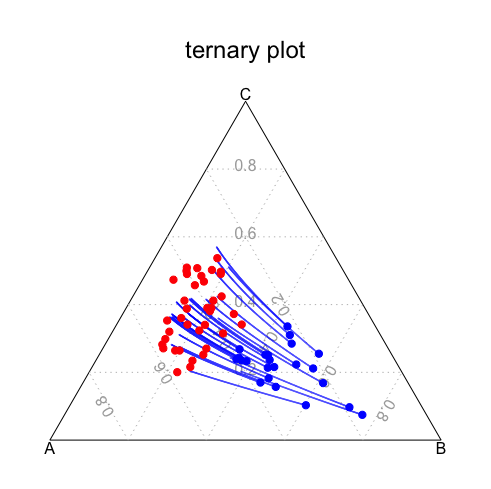}
    
{\small
\begin{tabular}{|c|p{0.065\linewidth}p{0.065\linewidth}p{0.075\linewidth}|}\hline
& ~~~A & ~~~B & ~~~C \\\hline
\textcolor{red}{$\bullet$} (\textcolor{red}{0}) & 43.713\% & 18.572\% & 37.715\% \\
\textcolor{blue}{$\bullet$} (\textcolor{blue}{1}) & 29.831\% & 48.605\% & 21.564\% \\
$T($\textcolor{red}{$\bullet$}$)$ & 29.553\% & 48.517\% & 21.930\% \\\hline 
\end{tabular}~~~~~~\begin{tabular}{|c|p{0.065\linewidth}p{0.065\linewidth}p{0.075\linewidth}|}\hline
& ~~~A & ~~~B & ~~~C \\\hline
\textcolor{red}{$\bullet$} (\textcolor{red}{0}) & 43.713\% & 18.572\% & 37.715\% \\
\textcolor{blue}{$\bullet$} (\textcolor{blue}{1}) & 29.831\% & 48.605\% & 21.564\% \\
$T($\textcolor{blue}{$\bullet$}$)$ & 43.728\% & 18.553\% & 37.719\% \\\hline 
\end{tabular}
}
\includegraphics[width=0.5\linewidth]{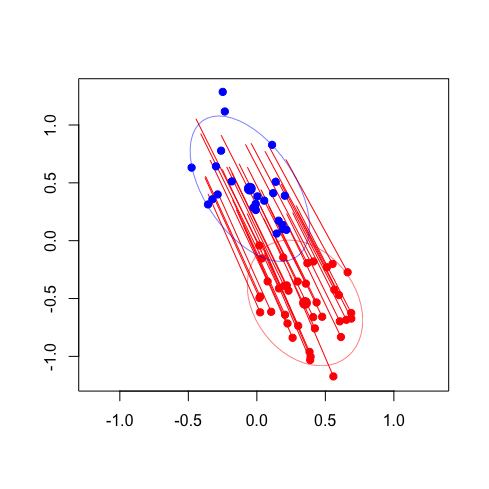}\includegraphics[width=0.5\linewidth]{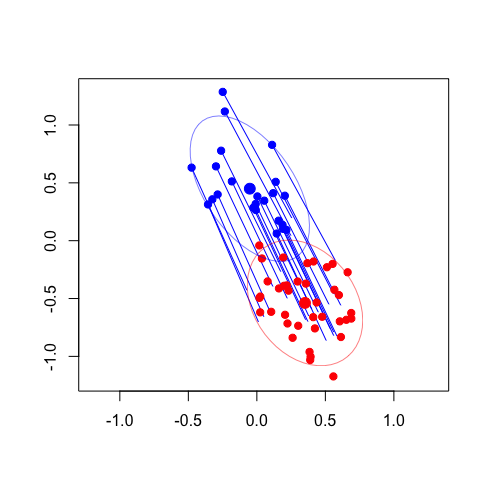}
    \caption{Counterfactuals using the $\operatorname{ilr}$ transformation, and Gaussian optimal transports, $\mu_{\textcolor{red}{0}}\mapsto\mu_{\textcolor{blue}{1}}$ on the left, and $\mu_{\textcolor{blue}{1}}\mapsto\mu_{\textcolor{red}{0}}$ on the right. Below are the averages of $\mathbf{x}_{\textcolor{red}{0},i}$'s and $\mathbf{x}_{\textcolor{blue}{1},i}$'s, and of the transported points. The lines are geodesics in the dual spaces, mapped in the simplex. Optimal transport in $\mathbb{R}^2$, on $\mathbf{z}_{\textcolor{red}{0},i}$'s and $\mathbf{z}_{\textcolor{blue}{1},i}$'s, can be visualized at the bottom (with linear mapping since Gaussian assumptions are made).} 
    \label{fig:ternary2}
\end{figure}

\subsection{Gaussian Mapping in the Euclidean Representation}\label{subsec:gaussian}

Given a random vector $\mathbf{X}$ in $\mathcal{S}_d$, we say
that $\mathbf{x}$ follows a ``normal distribution on the simplex'' if, for some isomorphism $h$, the vector of orthonormal coordinates, $\mathbf{Z}=h(\mathbf{X})$ follows a multivariate normal distribution on $\mathbb{R}^{d-1}$. If we suppose that both $\mathbf{X}_0$ and $\mathbf{X}_1$, taking values in $\mathcal{S}_d$, follow ``normal distributions on the simplex,'' then we can use standard Gaussian optimal transport, between $\mathbf{Z}_0$ and $\mathbf{Z}_1$. For convenience, suppose that the same isomorphism is used for both distributions (but that assumption can easily be relaxed). Hence, if \(\mathbf{Z}_0\sim\mathcal{N}(\boldsymbol{\mu}_0,\boldsymbol{\Sigma}_0)\) and \(\mathbf{Z}_1\sim\mathcal{N}(\boldsymbol{\mu}_1,\boldsymbol{\Sigma}_1)\), the optimal mapping is  linear,
\begin{equation}\label{eq:t}
\mathbf{z}_{1} = T^\star(\mathbf{z}_{0})=\boldsymbol{\mu}_{1} + \boldsymbol{A}(\mathbf{z}_{0}-\boldsymbol{\mu}_{0}),    
\end{equation}
where \(\boldsymbol{A}\) is a symmetric positive matrix that satisfies \(\boldsymbol{A}\boldsymbol{\Sigma}_{0}\boldsymbol{A}=\boldsymbol{\Sigma}_{1}\), which has a unique solution given by \(\boldsymbol{A}=\boldsymbol{\Sigma}_{0}^{-1/2}\big(\boldsymbol{\Sigma}_{0}^{1/2}\boldsymbol{\Sigma}_{1}\boldsymbol{\Sigma}_{0}^{1/2}\big)^{1/2}\boldsymbol{\Sigma}_{0}^{-1/2}\), where \(\boldsymbol{M}^{1/2}\) is the square root of the square (symmetric) positive matrix \(\boldsymbol{M}\) based on the Schur decomposition (\(\boldsymbol{M}^{1/2}\) is a positive symmetric matrix), as described in \cite{higham2008functions}. Interestingly, it is possible to derive McCann's displacement interpolation, from \cite{mccann1997convexity}, to have some sort of continuous mapping $T_t^\star$ such that $T_1^\star=T^\star$ and $T_0=Id$, and so that $\mathbf{Z}_{t}=T_t^\star(\mathbf{Z}_{0})$ has distribution \(\mathcal{N}(\boldsymbol{\mu}_t,\boldsymbol{\Sigma}_t)\) where \(\boldsymbol{\mu}_t=(1-t)\boldsymbol{\mu}_0+t\boldsymbol{\mu}_1\) and 
$$
\boldsymbol{\Sigma}_t = \boldsymbol{\Sigma}_0^{-1/2} \left( (1 - t) \boldsymbol{\Sigma}_0 + t \left( \boldsymbol{\Sigma}_0^{1/2} \boldsymbol{\Sigma}_1 \boldsymbol{\Sigma}_0^{1/2} \right)^{1/2} \right)^2 \boldsymbol{\Sigma}_0^{-1/2}.
$$

\begin{algorithm}[tb]
\caption{Gaussian Based Transport of $\mathbf{x}_0$ on $\mathcal{S}_d$}\label{alg:1}
\textbf{Input}: $\mathbf{x}_0$ ($\in\mathcal{S}_d$)\\
    \textbf{Parameter}: $\{\mathbf{x}_{0,1},\cdots,\mathbf{x}_{0,n_0}\}$ and $\{\mathbf{x}_{1,1},\cdots,\mathbf{x}_{1,n_1}\}$ in $\mathcal{S}_d$;\\
    \phantom{\textbf{Parameter}:} isomorphic transformation $h:\mathcal{S}_d\to\mathbb{R}^{d-1}$\\
    \textbf{Output}: $\mathbf{x}_{1}$
\begin{algorithmic}
\FOR{\(i\in\{1,\cdots,n_0\}\)}
\STATE $\mathbf{z}_{0,i}\gets h(\mathbf{x}_{0,i})$
\ENDFOR\\
\FOR{\(i\in\{1,\cdots,n_1\}\)}
\STATE $\mathbf{z}_{1,i}\gets h(\mathbf{x}_{1,i})$
\ENDFOR\\
\STATE $\mathbf{m}_0\gets$ average of $\{\mathbf{z}_{0,1},\cdots,\mathbf{z}_{0,n_0}\}$\\
\STATE $\mathbf{m}_1\gets$ average of $\{\mathbf{z}_{1,1},\cdots,\mathbf{z}_{1,n_1}\}$\\
\STATE $\mathbf{S}_0\gets$ empirical variance matrix of $\{\mathbf{z}_{0,1},\cdots,\mathbf{z}_{0,n_0}\}$\\
\STATE $\mathbf{S}_1\gets$ empirical variance matrix of $\{\mathbf{z}_{1,1},\cdots,\mathbf{z}_{1,n_1}\}$\\
\STATE $\boldsymbol{A}\gets \boldsymbol{S}_{0}^{-1/2}\big(\boldsymbol{S}_{0}^{1/2}\boldsymbol{S}_{1}\boldsymbol{S}_{0}^{1/2}\big)^{1/2}\boldsymbol{S}_{0}^{-1/2}$\\
\STATE $\mathbf{x}_{1} \gets \displaystyle{h^{-1}\big(\mathbf{m}_{1} + \boldsymbol{A}(h(\mathbf{x}_{0})-\mathbf{m}_{0})\big)}$\\
\end{algorithmic}
\end{algorithm}

Empirically, this can be performed using Algorithm \ref{alg:1}, and a simulation can be visualized in Figure \ref{fig:ternary2}, where $h=\operatorname{clr}$. On the left, we can visualize the mapping of red points to the blue distribution, and on the right, the ``inverse mapping" of blue points to the red distribution. Transformed points $\mathbf{z}=h(\mathbf{x})$, that are plotted at the bottom, are supposed to be normally distributed, and a {multivariate} Gaussian Optimal Transport mapping is used. Hence, $T_t^\star$ is linear in $\mathbb{R}^{d-1}$, as given by expression \ref{eq:t}, as well as displacement interpolation, corresponding to red and blue segments. But, as we can see on top of Figure \ref{fig:ternary2}, in the original space, $t\mapsto \mathbf{x}_{t}:=h^{-1}(\mathbf{z}_{t})$ will be nonlinear. Tables are average values of the three components of $\mathbf{x}$'s and $T^\star(\mathbf{x})$'s.

\section{Optimal Transport for Measures on $\mathcal{S}_d$}\label{sec:Sd:OT}

\subsection{Theoretical Properties}\label{sec:theory}

A function $\psi:\mathcal{S}_d\to\mathbb{R}$ is exponentially concave if $\exp[\psi]:\mathcal{S}_d\to\mathbb{R}_+$ is concave. As a consequence, such a function $\psi$ is differentiable almost everywhere. Let $\nabla \psi$ and $\nabla_{\Vec{u}} \psi$ denote, respectively, its gradient, and its directional derivative. Following \cite{pal2016geometry,pal2018exponentially,pal2020multiplicative}, define an allocation map generated by $\psi$, $\pi_\psi:\mathcal{S}_d\to\mathcal{S}_d$ defined as 
$$
\pi_\psi(\mathbf{x})=\left[
x_1\big(1+\nabla_{\Vec{e}_1-\mathbf{x}} \psi(\mathbf{x}) \big),
\cdots,
x_d\big(1+\nabla_{\Vec{e}_d-\mathbf{x}} \psi(\mathbf{x}) \big)
\right],
$$
where $\{\Vec{e}_1,\cdots,\Vec{e}_{d}\}$ is the standard orthonormal basis of $\mathbb{R} ^{d}$. 
Consider the optimal transport problem with the following cost function, on $\mathcal{S}_d\times\mathcal{S}_d$, i.e., the L-divergence corresponding to the cross-entropy,
\begin{equation}\label{eq:cost}
 c(\mathbf{x},\mathbf{y})=\log\left(\frac{1}{d}\sum_{i=1}^d\frac{y_i}{x_i}\right)-\frac{1}{d}\sum_{i=1}^d\log\left(\frac{y_i}{x_i}\right),
\end{equation}
called ``Dirichlet transport'' in \cite{baxendale2022random}. See \cite{pistone2024unified} for a discussion about the connections with the distance induced by Aitchsion's inner product of Equation (\ref{eq:inner}).
From Theorem~1 in \cite{pal2020multiplicative}, for this cost function, there exists an exponentially concave function $\psi^\star:\mathcal{S}_d\to\mathbb{R}$ such that 
$$
T^\star(\mathbf{x}) = \mathbf{x}\diamond \pi_{\psi^\star}\big(\mathbf{x}^{-1}\big)
$$
defines a push-forward from $\mathbb{P}_0$ to $\mathbb{P}_1$, and the coupling $(\mathbf{x},T^\star(\mathbf{x}))$ is optimal for problem (\ref{eq:transport:1}), and is unique if $\mathbb{P}_0$ is absolutely continuous. Observe that if $\mathbf{y}=T^\star(\mathbf{x})$,
$$
\mathbf{y} = \mathcal{C}\big(\pi_{\psi^\star}\big(\mathbf{z}\big)_1/z_1,\cdots,\pi_{\psi^\star}\big(\mathbf{z}\big)_d/z_d\big),$$
where $\mathbf{z}=\mathbf{x}^{-1}$.

One can also consider an interpolation, 
$$
T^\star_t(\mathbf{x}) = \mathbf{x}\diamond \pi_{t}\big(\mathbf{x}^{-1}\big)
$$
where $\pi_{t}=(1-t)d^{-1}\boldsymbol{1}+t\pi_{\psi^\star}$ (even if this approach differs from McCann's displacement interpolation).

\begin{figure}[!htb]
    \centering
    \includegraphics[width=0.5\linewidth]{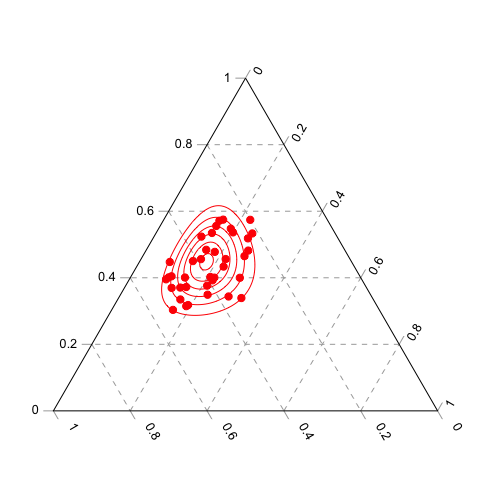}\includegraphics[width=0.5\linewidth]{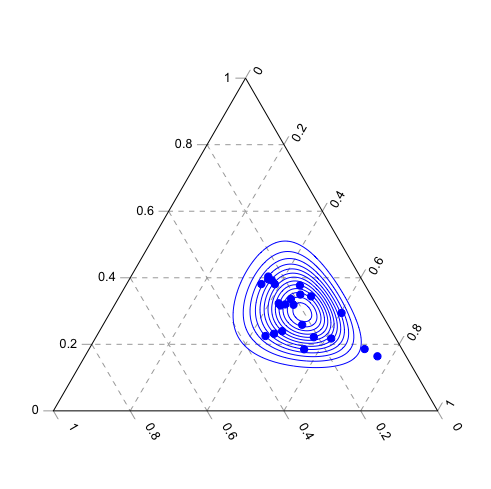}
    \caption{Densities of Dirchlet distributions in $\mathcal{S}_3$ fitted on observations of the toy dataset of Figure~\ref{fig:ternary1}.}
    \label{fig:ternary3}
\end{figure}

Note that a classical distribution on $\mathcal{S}_d$ is Dirichlet distribution, with density
\[{\displaystyle f\left(x_{1},\ldots ,x_{d};\boldsymbol {\alpha }\right)={\frac {1}{\mathrm {B} ({\boldsymbol {\alpha }})}}\prod _{i=1}^{d}x_{i}^{\alpha _{i}-1}}\]
for some ${\boldsymbol {\alpha }}=(\alpha _{1},\ldots ,\alpha _{d})\in\mathbb{R}_+^d$, 
and a normalizing constant denoted $\mathrm {B} ({\boldsymbol {\alpha }})$. Level curves of the density of Dirichlet distributions fitted on our toy dataset can be visualized in Figure \ref{fig:ternary3}.
Unfortunately, unlike the multivariate Gaussian distribution, there is no explicit expression for the optimal mapping between Dirichlet distribution (regardless of the cost). Therefore, to remain within $\mathcal{S}_d$ and avoid the $\mathbb{R}^{d-1}$ representation, numerical techniques should be considered.

\subsection{Matching}\label{subsec:matching}

Consider two samples in the $\mathcal{S}_d$ simplex, $\{\mathbf{x}_{0,1},\cdots,\mathbf{x}_{0,n_0}\}$ and $\{\mathbf{x}_{1,1},\cdots,\mathbf{x}_{1,n_1}\}$. The discrete version of the {Kantorovich} problem (corresponding to Equation \ref{eq:transport:2}) is
\begin{equation}\label{prog:match:2}
\underset{\mathrm{P}\in   U(n_{0},n_{1})}{\min} \left\lbrace \sum_{i=1}^{{n_{0}}} \sum_{j=1}^{{n_{1}}} \mathrm{P}_{i,j}\mathrm{C}_{i,j} \right\rbrace
\end{equation}
where, as in \cite{brualdi2006combinatorial}, $ U(n_{0},n_{1})$ is the set of ${n_{0}}\times {n_{1}}$ matrices corresponding to the {convex transportation polytope} 
$$
 U(n_{0},n_{1})=\left\lbrace
 \mathrm{P}:\mathrm{P}\boldsymbol{1}_{{n_{1}}}=\boldsymbol{1}_{n_{0}}\text{ and }{\mathrm{P}}^\top\boldsymbol{1}_{{n_{0}}}=\frac{n_0}{n_1}\boldsymbol{1}_{n_{1}}
 \right\rbrace,
 $$
 and where $\mathrm{C}$ denotes the ${n_{0}}\times {n_{1}}$ cost matrix, $\mathrm{C}_{i,j}=c(\mathbf{x}_i,\mathbf{x}_{j})$, associated with cost from Equation (\ref{eq:cost}).

In Algorithm \ref{alg:2}, we recall how this procedure works, which is the one explained in \cite{peyre2019computational}, with a specific cost function (from Equation (\ref{eq:cost})). In the toy dataset, this can be visualized for two specific observations $\mathbf{x}_{0,i}$. If $n_0\neq n_1$, it is not a one-to-one coupling, and ``the counterfactual'' is actually a weighted average of $\mathbf{x}_{1,j}$'s, where weights are given in row $\mathbf{P}^\star_i=[\mathbf{P}^\star_{i,1},\cdots,\mathbf{P}^\star_{i,n_1}]\in\mathcal{S}_{n_1}$.

\begin{algorithm}[tb]
\caption{Coupling samples on $\mathcal{S}_d$}\label{alg:2}
\textbf{Input}: $\{\mathbf{x}_{0,1},\cdots,\mathbf{x}_{0,n_0}\}$ and $\{\mathbf{x}_{1,1},\cdots,\mathbf{x}_{1,n_1}\}$ in $\mathcal{S}_d$;\\
\textbf{Output}: weight matching matrix $n_0\times n_1$ $\mathbf{P}^*$\\
\vspace{-.4cm}
\begin{algorithmic}
    \STATE  $\mathbf{C} \gets $ matrix $n_0\times n_1$, $\mathbf{C}_{i,j}=c(\mathbf{x}_i,\mathbf{x}_j)$ using (\ref{eq:cost})\\
    \STATE  $\mathbf{P}^\star \gets$ solution of Equation (\ref{prog:match:2}), using LP libraries
\end{algorithmic}
\end{algorithm}

\begin{figure}[!htb]
    \centering
    \includegraphics[width=0.5\linewidth]{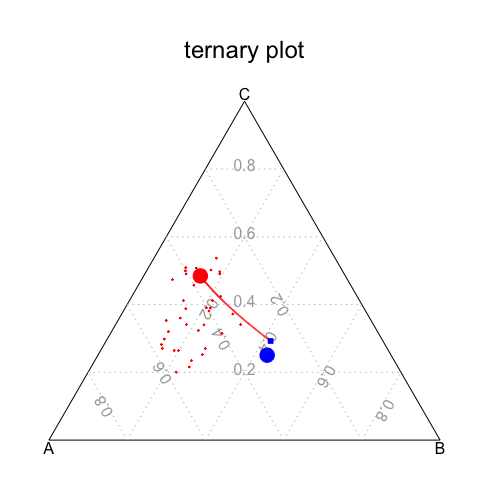}\includegraphics[width=0.5\linewidth]{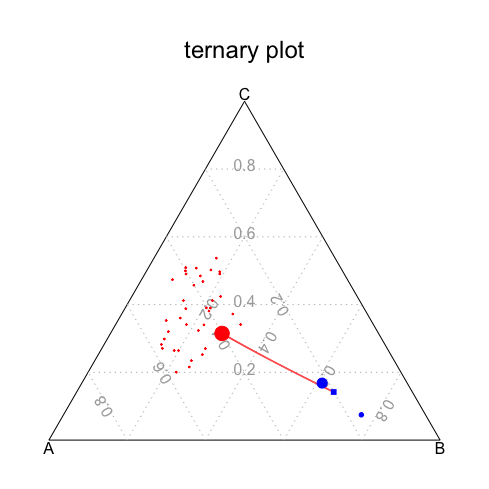}
\caption{Getting empirical counterfactuals using matching techniques, with $\mathbf{x}_{\textcolor{red}{0},i}$ in \textcolor{red}{red} (on the top-left hand-side), and counterfactuals  $\mathbf{x}_{\textcolor{blue}{1},j}$'s in \textcolor{blue}{blue} (bottom-right hand-side), with size proportional to $\mathbf{P}^\star_i=[\mathbf{P}^\star_{i,1},\cdots,\mathbf{P}^\star_{i,n_1}]\in\mathcal{S}_{n_1}$.}\label{fig:coupling}
\end{figure}

\section{Application on Sequential Transport for Counterfactuals}\label{sec:application}

Variables $\mathbf{x}_j$ in tabular data are either continuous or categorical. If $\mathbf{x}_j$ is continuous, since $\mathbf{x}_j\in\mathbb{R}$, transporting from observed $\mathbf{x}_j|s=0$ to counterfactual $\mathbf{x}_j|s=1$ is performed using standard (conditional) monotonic mapping, as discussed in \cite{machado2024sequential}, using classical $F_1^{-1}\circ F_0$. If $\mathbf{x}_j$ is categorical, with $d$ categories, consider some fitted model $\widehat{m}(\mathbf{x}_j|\mathbf{x}_{-j})$, using some multinomial loss, and let $\widehat{\mathbf{x}}_j=\widehat{m}(\mathbf{x}_j|\mathbf{x}_{-j})$ denote the predicted scores, so that $\widehat{\mathbf{x}}_j\in\mathcal{S}_d$. Then use Algorithm \ref{alg:1}, with a Gaussian mapping in an Euclidean representation space, to transport from observed $\widehat{\mathbf{x}}_j|s=0$ to counterfactual $\widehat{\mathbf{x}}_j|s=1$, in $\mathcal{S}_d$.

\subsection{German Credit: Purpose}\label{sec:germancredit}

In the popular \texttt{German Credit} dataset, from \cite{misc_german}, the variable \texttt{Purpose} described the reason an individual took out a loan. This variable is an important predictor for explaining potential defaults. The original variable is based on ten categories, that are merged here into three main classes, {\footnotesize\sffamily cars}, {\footnotesize\sffamily equipment} and {\footnotesize\sffamily other}, in order to visualize the transport in a ternary plot --- or Gibbs triangle. The sensitive variable $s$ is here \texttt{Sex}.

\begin{figure}[!htb]
    \centering
    \includegraphics[width=0.5\linewidth]{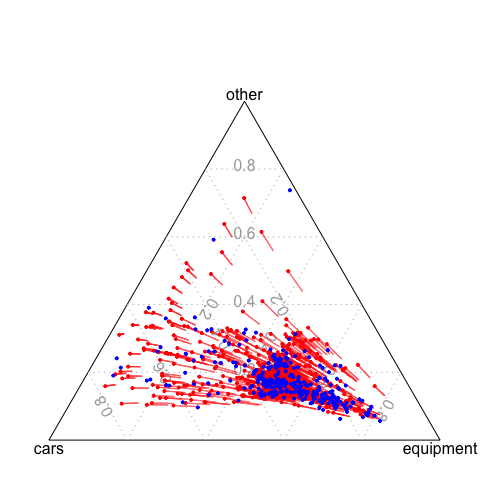}\includegraphics[width=0.5\linewidth]{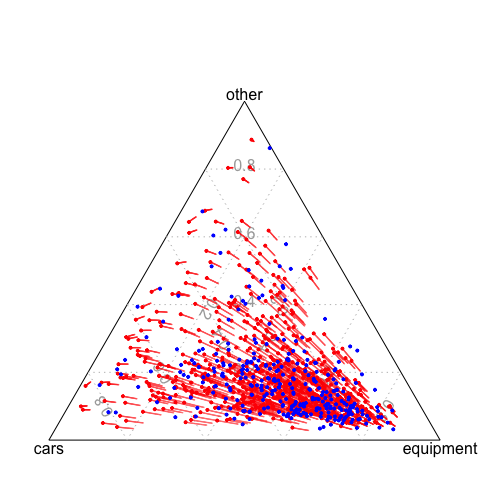}
    
{\small
\begin{tabular}{|c|}
\multicolumn{1}{c}{}  \\
\multicolumn{1}{c}{}  \\\hline
categorical \textcolor{blue}{$\bullet$} (M) \\
categorical \textcolor{red}{$\bullet$} (F)  \\\hdashline[1pt/1pt]
composition \textcolor{blue}{$\bullet$} (M) \\
composition \textcolor{red}{$\bullet$} (F)  \\
$T($\textcolor{red}{$\bullet$}$)$ \\\hline 
\end{tabular}~~~\begin{tabular}{|p{0.065\linewidth}p{0.065\linewidth}p{0.075\linewidth}|}
\multicolumn{3}{c}{MLR (1)}  \\\hline
 ~~~~cars & equipmt. & ~~other \\\hline
 30.323\%&    53.226\%&    16.452\% \\
 35.217\%&   44.638\%&    20.145\%  \\\hdashline[1pt/1pt]
 31.106\%&    51.328\%&    17.565\% \\
 34.865\%&     45.490\%&    19.645\% \\
 31.016\%& 51.418\%& 17.566\% \\\hline 
\end{tabular}~~~\begin{tabular}{|p{0.065\linewidth}p{0.065\linewidth}p{0.075\linewidth}|}
\multicolumn{3}{c}{MLR (2)}  \\\hline
 ~~~~cars & equipmt. & ~~other \\\hline
 30.323\%&    53.226\%&    16.452\% \\
 35.217\%&   44.638\%&    20.145\%  \\\hdashline[1pt/1pt]
 31.955\%&    50.539\%&    17.507\% \\
 34.484\%&    45.845\%&    19.671\% \\
 31.855\%& 50.570\%& 17.575\% \\\hline 
\end{tabular}
}
    \includegraphics[width=0.5\linewidth]{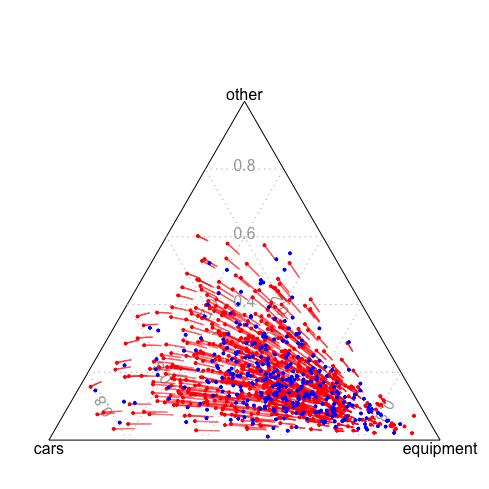}\includegraphics[width=0.5\linewidth]{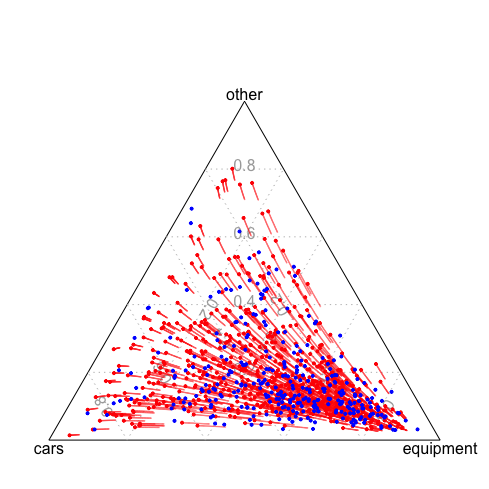}
    
{\small
\begin{tabular}{|c|}
\multicolumn{1}{c}{}  \\
\multicolumn{1}{c}{}  \\\hline
composition \textcolor{blue}{$\bullet$} (M) \\
composition \textcolor{red}{$\bullet$} (F)  \\
$T($\textcolor{red}{$\bullet$}$)$ \\\hline 
\end{tabular}~~~\begin{tabular}{|p{0.065\linewidth}p{0.065\linewidth}p{0.075\linewidth}|}\multicolumn{3}{c}{random forest}  \\\hline
~~~~cars & equipmt. & ~~other  \\\hline
 32.055\%&    49.882\%&    18.063\% \\
 34.977\%&    45.409\%&    19.614\% \\
 32.046\%& 49.883\%& 18.070\% \\\hline 
\end{tabular}~~~\begin{tabular}{|p{0.065\linewidth}p{0.065\linewidth}p{0.075\linewidth}|}\multicolumn{3}{c}{boosting}  \\\hline
~~~~cars & equipmt. & ~~other \\\hline
 31.843\%&     50.712\%&     17.445\% \\
 34.714\%&     45.341\%&     19.945\% \\
 31.871\%&  50.675\%&  17.455\% \\\hline 
\end{tabular}
}
    \caption{Optimal transport using the $\operatorname{clr}$ transformation, and Gaussian optimal transports, on the \texttt{purpose} scores in the \texttt{German Credit} database, with two logistic GAM models to predict scores, on top, and below a random forest (left) and a boosting model (right). Points in red are compositions for women, while points in blue are for men. Lines indicate the displacement interpolation when generating counterfactuals.} 
    \label{fig:ternary:german:1}
\end{figure}

{We aim to construct a counterfactual value for the loan purpose, assuming the individuals were of a different sex. To achieve this, we apply our suggested procedure from to represent the \texttt{purpose} categorical variable as a compositional variable, using the same four models outlined in Section~\ref{sec:categorical} and then apply Gaussian mapping from Section~\ref{subsec:gaussian}. The results provided by all of the models, shown in Figure~\ref{fig:ternary:german:1}, suggest that, had the individuals been of a different sex, the purpose of the loan would have changed. Specifically, if the average scores in each group ({\footnotesize\sffamily cars}, {\footnotesize\sffamily equipment}, and {\footnotesize\sffamily other}) were approximately $[35\%, 45\%, 20\%]$ in the female population, after transporting to obtain the counterfactuals, the average scores become $[31\%, 52\%, 18\%]$, which closely resemble the actual frequencies of each category in the original male population.}

\begin{figure}[!htb]
    \centering
    \includegraphics[width=0.5\linewidth]{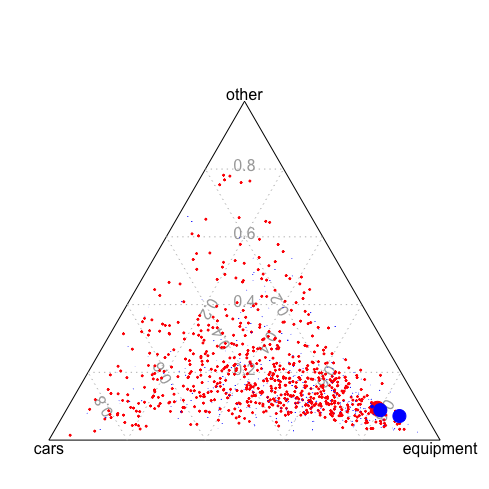}\includegraphics[width=0.5\linewidth]{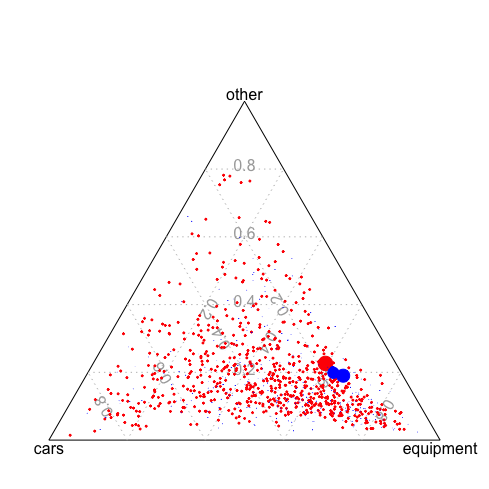} 
\caption{Empirical matching of two women, in red, from the \texttt{German Credit} dataset, with 2 or 3 men, in blue. Size of blue dots are proportional to the weights $\mathbf{P}^\star_i$.}\label{fig:conterf}
    \end{figure}

One can also consider our second approach, using matching in $\mathcal{S}_3$. Consider individual $i$ among women, e.g., the left of Figure \ref{fig:conterf}, 
$x_{0,i}=$ ``{\footnotesize\sffamily equipment}.'' Using a MLR model, we obtain composition $\mathbf{x}_{0,i}$, here $[11.38\%,79.30\%,9.32]$. Using Algorithm \ref{alg:2}, three points $\mathbf{x}_{1,j}$'s are matched, respectively with weights
$[0.453,0.094,0.453]$. The first and the third individuals are such that $x_{1,j}=$ ``{\footnotesize\sffamily equipment}'' too, the second one ``{\footnotesize\sffamily other}.'' So it would make sense to suppose that the counterfactual version of woman $i$ with an ``{\footnotesize\sffamily equipment}'' credit is a man with the same purpose. Actually, using Gaussian transport, $T^\star(\mathbf{x}_{0,i})=[15.78\%,     69.54\%,     14.68\%]$.

\subsection{Adult: Marital Status}

\begin{figure}[!htb]
    \centering
    \includegraphics[width=0.5\linewidth]{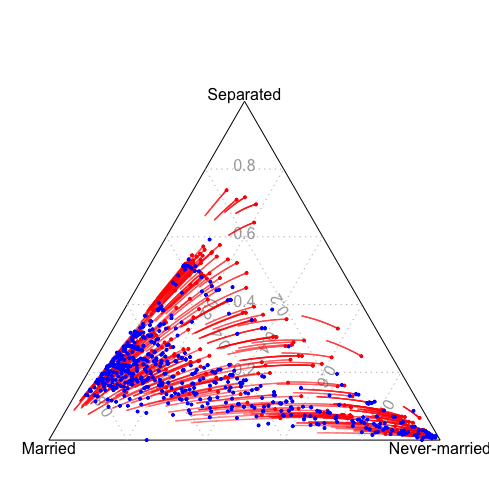}\includegraphics[width=0.5\linewidth]{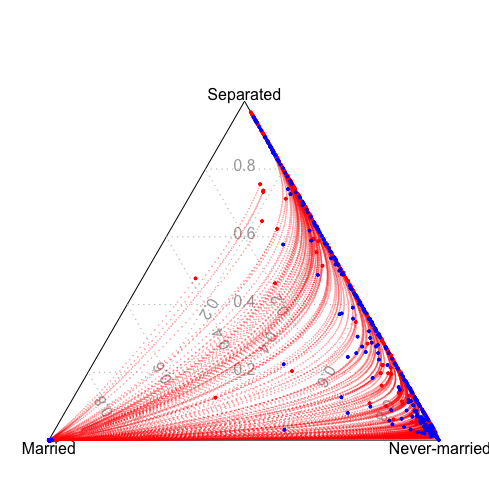}
      
{\small
\begin{tabular}{|c|}
\multicolumn{1}{c}{}\\
\multicolumn{1}{c}{}  \\\hline
categorical \textcolor{blue}{$\bullet$} (M) \\
categorical \textcolor{red}{$\bullet$} (F)  \\\hdashline[1pt/1pt]
composition \textcolor{blue}{$\bullet$} (M) \\
composition \textcolor{red}{$\bullet$} (F)  \\
$T($\textcolor{red}{$\bullet$}$)$ \\\hline 
\end{tabular}~~~\begin{tabular}{|p{0.065\linewidth}p{0.065\linewidth}p{0.075\linewidth}|}
\multicolumn{3}{c}{GAM-MLR (1)}  \\\hline
 ~~~Mar. & Never~M & ~~~Sep. \\\hline
  61.752\%&        26.609\%&        11.639\% \\
 15.130\%&        44.204\%&   40.667\%  \\\hdashline[1pt/1pt]
51.441\%&        29.641\%&        18.918\% \\
 35.180\%&        39.714\%&        25.106\% \\
 49.857\%& 30.740\%& 19.403\% \\\hline 
\end{tabular}~~~\begin{tabular}{|p{0.065\linewidth}p{0.065\linewidth}p{0.075\linewidth}|}\multicolumn{3}{c}{GAM-MLR (2)}  \\\hline
 ~~~Mar. & Never~M & ~~~Sep. \\\hline
 61.752\%&        26.609\%&        11.639\% \\
 15.130\%&        44.204\%&   40.667\%  \\\hdashline[1pt/1pt]
 59.819\%&        27.276\%&        12.905\% \\
 12.982\%&        46.093\%&        40.925\% \\
 67.214\%& 24.678\%&  8.108\% \\\hline 
\end{tabular}
}

    \includegraphics[width=0.5\linewidth]{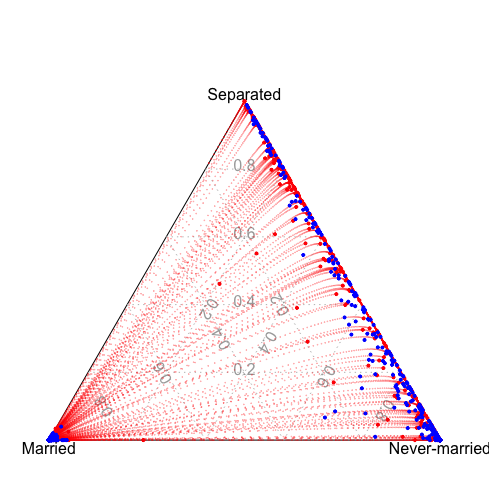}\includegraphics[width=0.5\linewidth]{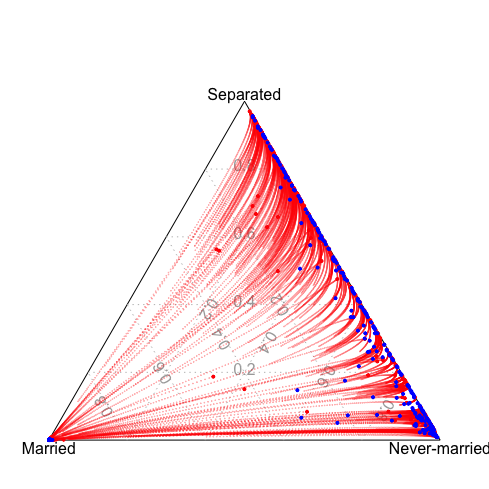}
      
{\small
\begin{tabular}{|c|}
\multicolumn{1}{c}{}  \\
\multicolumn{1}{c}{}  \\\hline
composition \textcolor{blue}{$\bullet$} (M) \\
composition \textcolor{red}{$\bullet$} (F)  \\
$T($\textcolor{red}{$\bullet$}$)$ \\\hline 
\end{tabular}~~~\begin{tabular}{|p{0.065\linewidth}p{0.065\linewidth}p{0.075\linewidth}|}\multicolumn{3}{c}{random forest}  \\\hline
 ~~~Mar. & Never~M & ~~~Sep. \\\hline
59.619\%&        27.827\%&        12.554\% \\
 12.765\%&        48.707\%&        38.528\% \\
 46.306\%& 34.967\%& 18.727\% \\\hline 
\end{tabular}~~~\begin{tabular}{|p{0.065\linewidth}p{0.065\linewidth}p{0.075\linewidth}|}\multicolumn{3}{c}{boosting}  \\\hline
 ~~~Mar. & Never~M & ~~~Sep. \\\hline
 59.764\%&        27.250\%&        12.987\% \\
 13.219\%&        45.976\%&        40.805\% \\
 47.611\%& 30.313\%& 22.076\% \\\hline 
\end{tabular}
}

 \caption{Optimal transport using the $\operatorname{clr}$ transformation, and Gaussian optimal transports, on the \texttt{Marital Status} scores in the \texttt{Adult} database, with two logistic GAM-MLR models to predict scores, on top, and below a random forest (left) and a boosting model (right). Points in red are compositions for women, while points in blue are for men. Lines indicate the displacement interpolation when generating counterfactuals.} \label{fig:adult}
\end{figure}

Following the numerical applications in \cite{plevcko2021fairadapt} and \cite{machado2024sequential}, we consider here the \texttt{Adult} dataset, from \cite{misc_adult_2}. We regrouped categories of the \texttt{Marital Status} variable to create three generic ones (that can be visualized in a ternary plot, as in Figure \ref{fig:adult}), namely {\footnotesize\sffamily Married}, {\footnotesize\sffamily Never-married} and {\footnotesize\sffamily Separated}. This example is interesting because if we compare status with respect to the \texttt{Sex} variable, proportions are quite different. In the dataset, proportions for married, never married, and separated are (roughly) $[62\%, 27\%, 12\%]$ for men, $[14\%, 44\%, 41\%]$ for women (more precise values are at the top of the table in Figure~\ref{fig:adult}). Thus, the counterfactual of a ``separated'' woman is more likely to be a ``married'' man than a ``separated'' man. Four models are used to convert the categorical variable \texttt{Marital Status} into a composition, as previously. The first MLR is based on three variables: a categorical variable, \texttt{occupation}, and two continuous ones, \texttt{age}, and \texttt{hours\_per\_week}, modeled nonlinearly using $b$-splines (hence, it is referred to as a logistic GAM). This model is clearly underfitted. Therefore, observations $\mathbf{x}_{\textcolor{red}{0},i}$'s for women and $\mathbf{x}_{\textcolor{blue}{1},i}$'s for men clearly are in the interior of $\mathcal{S}_d$. In contrast, the more complex MLR ({which uses additional features}), 
as well as the random forest and boosting models, can produce predictions near the simplex boundary, $\partial\mathcal{S}_d$.

For the underfitted model (top left), transported scores have a distribution very close to the ones in the population of men. For the more accurate MLR model (top right), proportions are very close to the actual proportions (which is not surprising since GLMs are usually well calibrated), but the transported scores are slightly different than the proportions of categories (proportions were $[62\%, 27\%, 12\%]$ while average transported scores are $[67\%, 25\%, 8\%]$). At least, we are different from the original ones, but the mapping is not as accurate as it should be. This might come from the fact that when the points $\mathbf{x}_i$ are close to the border $\partial\mathcal{S}_d$, it is quite unlikely that the sample $\mathbf{z}_i$ is Gaussian.

\section{Conclusion}

{In this article, we introduce a novel approach for constructing counterfactuals for categorical data by transforming them into compositional data using a probabilistic classifier. Our approach avoids imposing arbitrary assumptions about label ordering. However, our methodology is not without limitations. OT computations, particularly on the simplex, can be computationally intensive for large-scale datasets, posing challenges in high-dimensional settings. Additionally, the reliance on a probabilistic classifier in the initial step introduces potential vulnerabilities. Biases may arise from a poorly calibrated or inaccurate classifier, impacting the quality of the subsequent analysis---especially with scarce categories that may need grouping to apply the proposed method.}

\appendix


\section{Complexity of the Main Algorithms}

The complexity of Algorithm~\ref{alg:1} is $\mathcal{O}(d^3\!+\!(n_0\!+\!n_1) d^2)$ primarily arising from the computation of the class-wise covariance matrices $S_0$, $S_1$, and the transformation matrix $A$. When the number of classes $d$ becomes large, this cost becomes prohibitive; in such scenarios, alternative OT formulations (beyond the Gaussian OT mapping) are advisable.

Alternatively, Algorithm~\ref{alg:2} provides a more scalable approach in high-dimensional settings. Here, $d$ only affects the cost matrix computation with complexity $\mathcal{O}(n_0 n_1 d)$, while the dominant computational burden lies in solving the OT problem, which scales as $\mathcal{O}((n_0 n_1)^{3/2})$ when using the Operator Splitting Quadratic Program (OSQP) solver.

\section{Choice of the Cost Function in Algorithm \ref{alg:2}}

Section~\ref{sec:Sd:OT} introduces an approach to transporting probability measures directly on the simplex $\mathcal{S}_d$. More precisely, Section~\ref{sec:theory} defines the existence of an OT map $T^\star$ from a source measure $\mathbb{P}_0$ on $\mathcal{S}_d$ to a target measure $\mathbb{P}_1$ on $\mathcal{S}_d$, using the cross-entropy as the cost function, which is referred to as ``Dirichlet transport'' \citep{baxendale2022random}. While this cost function guarantees the theoretical existence of an optimal map, no closed-form expression for $T^\star$ is available in practice. Instead, numerical optimization is performed via the Kantorovich formulation of the OT problem to match individuals from $\mathbb{P}_0$ to $\mathbb{P}_1$ on $\mathcal{S}_d$, as described in Section \ref{subsec:matching}. Although Euclidean costs, such as the Wasserstein-1 or Wasserstein-2 distances, could be considered, they violate the geometry of the simplex by relying on absolute differences, unlike the cross-entropy, which accounts for relative proportions.

\section{Integration of the methodology into a SCM for counterfactual fairness assessment}\label{sec:appendix-scm}

In Section~\ref{sec:categorical}, to transform a categorical variable $x$ into a numerical one, we suggest using a probabilistic classifier based on the features $\mathcal{X}_{-x}$, i.e., all features except the categorical variable. In this appendix, we consider a more global approach, within an SCM, which requires positing a structural causal model in advance. Consider the DAG shown in Figure~\ref{fig:dag-example}, whose topological order is

$$
S \;\rightarrow\; X_1 \;\rightarrow\; X_2 \;\rightarrow\; X_3 \;\rightarrow\; Y,
$$

where $S\in\{0,1\}$ is the sensitive treatment (for example, binary gender), $X_1\in\mathbb{R}$ is a numeric feature, $X_2\in[\![d_2]\!]$ and $X_3\in[\![d_3]\!]$ are categorical features with $d_2$ and $d_3$ categories, respectively, and $Y$ is the outcome sink. To generate a counterfactual for an individual with $S=0$, we first flip the treatment by setting $S=1$. We then obtain the numeric counterfactual $X_1^\star$ using a chosen mapping mechanism (e.g.,  using optimal transport).

Subsequently, we construct each categorical counterfactual in sequence. For $X_2$, we fit a classifier on its parents $(S, X_1)$ to estimate the conditional distribution $\widehat P(X_2\mid S,X_1)$. This yields probability vectors in the simplex $\mathcal{S}_{d_2}$ for both the factual group $(S=0)$ and the counterfactual group $(S=1)$, which needs to be converted to a single label $X_2^\star$ in $\{1,\dots,d_2\}$. We suggest either sampling according to the probabilities or selecting the highest‐probability category (top-label). We repeat the same process for $X_3$, estimating $\widehat P(X_3\mid S,X_2)$ and mapping its output into $\{1,\dots,d_3\}$. Finally, we predict the counterfactual outcome $Y^\star$ with a model conditioned on $(S,X_1^\star,X_2^\star,X_3^\star)$. This sequential, topologically ordered procedure embeds each probabilistic result into its required discrete space, ensuring consistency with the \textit{a priori} SCM.

\begin{figure}[htb!]
    \centering
    \begin{tikzpicture}[
    >={Stealth},
    every node/.style={font=\sffamily},
    sens/.style   = {circle, draw, text=white, fill=wongGreen,  minimum size=8mm},
    var1/.style   = {circle, draw, text=white, fill=wongOrange, minimum size=8mm},
    var2/.style   = {circle, draw, text=white, fill=wongPurple, minimum size=8mm},
    output/.style    = {circle, draw, text=white, fill=gray,   minimum size=8mm},
    edge/.style   = {->, thick, black}
]

  \node[sens] (S)  at (0,0)        {S};
  \node[var1] (X1) at (2,1.5)      {$X_1$};
  \node[var2] (X2) at (2,0)        {$X_2$};
  \node[var2] (X3) at (2,-1.5)     {$X_3$};
  \node[output]  (Y)  at (4,0)        {$Y$};

  \draw[edge] (S) -- (X1);
  \draw[edge] (S) -- (X2);
  \draw[edge] (S) -- (X3);
  \draw[edge] (S) to[bend right=30] (Y);

  \draw[edge] (X1) -- (X2);
  \draw[edge] (X1) -- (Y);
  \draw[edge] (X2) -- (X3);
  \draw[edge] (X2) -- (Y);
  \draw[edge] (X3) -- (Y);

\end{tikzpicture}
    \caption{Example of a Structural Causal Model with a \textcolor{wongGreen}{sensitive attribute $S$}, a \textcolor{orange}{numeric variable $X_1$}, \textcolor{wongPurple}{two categorical variables $X_2$ and $X_3$}, and an \textcolor{gray}{output variable $Y$}.}
    \label{fig:dag-example}
\end{figure}

Matching individuals using their probability vectors on $\mathcal{S}_d$ allows to uniquely determine them as it involves continuous distributions on $[0,1]$ for each category. In contrast, if counterfactuals were computed directly from categorical data, one would need to rely on the non-deterministic counterfactual framework described in \cite[Chap. 4]{Pearl2016CausalII}.

\bibliography{biblio}

\end{document}